\documentclass[11pt]{article}

\usepackage[margin=1in]{geometry}
\usepackage[T1]{fontenc}
\usepackage[utf8]{inputenc}
\usepackage{lmodern}
\usepackage{amsmath,amssymb}
\usepackage{graphicx}
\usepackage{booktabs}
\usepackage{caption}
\usepackage[colorlinks,urlcolor=blue,linkcolor=blue,citecolor=blue]{hyperref}
\usepackage{xurl}   
\usepackage{algorithm}
\usepackage{algpseudocode}
\usepackage[protrusion=true,expansion=true]{microtype}

\graphicspath{{figures/}}

\newcommand{\MASK}{\texttt{[MASK]}}

\title{Solvable Sokoban Without a Solver via Diffusion}
\author{Sina Baghal \\ \texttt{\href{mailto:siinabaghal@gmail.com}{siinabaghal@gmail.com}}}
\date{}

\begin{document}
\maketitle

\begin{abstract}
\setlength{\parskip}{\medskipamount}
\setlength{\parindent}{0pt}
\noindent
Deciding whether a Sokoban puzzle is solvable is PSPACE-complete \cite{culberson1997sokoban}: solutions can
be exponentially long and there is no short certificate to check. Solvability is also a fragile property since even a single misplaced wall can silently render an entire puzzle unsolvable.

In this work, we show that a transformer-based discrete diffusion model
trained purely on tile completion, with no access to solvers, rewards, or
solvability labels, achieves a solvability rate of 77.4\%, with
94.5\% of the remaining failures rendered solvable by removing a single wall.
In other words, a global, search-heavy property follows from a local training
objective:
\begin{center}
\emph{Trained only to fill in masked cells, the model inherits solvability it was never trained on.}
\end{center}

An autoregressive model factorizes as $p(c_k \mid c_1 \dots c_{k-1})$ meaning a
fixed order and always conditioned on a prefix. Masked diffusion doesn't: it
hides a random subset of cells and learns $p(c_k \mid \text{any subset})$, so at
generation time it can reveal cells in any order, each one conditioned on
everything already placed wherever it sits on the board. A puzzle's difficulty comes from exactly this kind of non-local interaction: a
decision in one part of the grid constraining what will work somewhere else
entirely. As such, a generator that isn't locked into a single fixed order is a better
structural match for the problem than one that is.

The figure below follows one generated puzzle across a single sampling run,
from a fully masked grid to a finished board. The model only ever fills in
masked cells; the finished board is solvable.

\begin{center}
  \includegraphics[width=\linewidth]{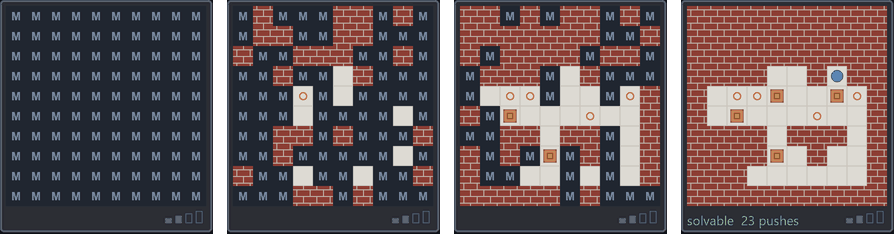}
\end{center}

The model's training pipeline is adapted from MD4 \cite{shi2024md4}, and the
dataset is DeepMind's Boxoban \cite{guez2019boxoban}. Both the trained model and
instructions for generating puzzles are publicly
\href{https://github.com/sinabaghal/SokobanPlayground}{available}.

\textbf{AI usage.} The research question, model architecture, training setup, and
all experimental design decisions are the author's; Claude Code was used as an
implementation and editing assistant. 
\end{abstract}

\newpage

\section{Preliminary}

This section is organized as follows. We first define the game of Sokoban, how a
puzzle's solvability is decided using a push-based solver, and how its
difficulty is measured. We then describe the masked diffusion model: its
mechanism, its evolution from the continuous-diffusion formulation, and its
suitability for Sokoban puzzle generation.

\subsection{Sokoban}

Sokoban is a single-player puzzle created in Japan around 1980. Played on a
grid-based maze with a single character and multiple boxes, the objective is to
move all boxes onto designated target positions. Game mechanics are strictly
restricted to pushes, meaning that the player can only push one box at a time
into an adjacent unoccupied space and cannot pull boxes.

Due to spatial interdependencies, Sokoban cannot be decomposed into isolated
tasks. Moving a single box alters board topology and player reachability; an
incorrect execution order can obstruct future paths or render previously placed
boxes into deadlocks. Sequence is therefore as vital as destination. Because
localized, step-by-step decision-making fails, players must formulate a holistic
plan accounting for all box interactions prior to execution.

\subsubsection{Solvability}

A puzzle is solvable if some sequence of legal box pushes lands every box on a
goal. We decide this with a push-based solver that branches on pushes, not on
player moves. The player's individual steps only relocate the worker without
changing the puzzle, so branching on every movement would blow up the search
with positions that differ solely in where the player stands. Instead we branch
once per box push, and normalize the player to the region it can currently
reach: every board with identical boxes and the player anywhere inside that
reachable region collapses to a single search state. Branching is therefore tied
to the box configuration rather than to navigation. We also prune dead cells:
working backwards from each goal by pulling a box outward, any cell never
reached is one no box could ever be pushed to a goal from, so pushes into it are
discarded rather than branched on.

Culberson \cite{culberson1997sokoban} proved that deciding Sokoban solvability
is PSPACE-complete. PSPACE is the class of problems solvable with a polynomial
amount of memory, though possibly requiring exponential time, and it contains
NP. What separates Sokoban from an NP-complete puzzle such as Sudoku is that its
shortest solution can be exponentially long: there is no short certificate that
a checker could verify quickly, so establishing that a puzzle is solvable may
require searching an enormous state space.

\subsubsection{Difficulty}

Jaru\v{s}ek and Pel\'anek \cite{jarusek2010difficulty} modeled Sokoban difficulty
on the push-based state-space graph $G = (V, E)$, where each vertex $v \in V$ is
a game state (box positions plus the player's reachable area) and each directed
edge $e = (u, v) \in E$ is a single valid box push. Evaluating metrics against
large-scale human solving logs, they found that static, global properties of the
graph fail to predict human difficulty: push-space size $|V|$ showed no
significant correlation ($r = -0.11$) and shortest push-solution length was
only weakly predictive ($r = 0.30$). What did predict difficulty were metrics
modelling the search a human actually performs, rather than properties of the
finished graph. Two stood out. A decomposition metric, measuring how far a
puzzle breaks into independent sub-problems that can be solved one at a time,
reached $\rho = 0.82$ under Spearman correlation. A stochastic model of a human
wandering the state space, rather than walking the optimal path, reached
$r = 0.76$ under Pearson correlation.

The difficulty measure used in this work is aligned with that finding. We rate a
puzzle by the number of states the push solver expands before it finds a
solution, which is a measure of how much search the puzzle demands rather than
of how large its state space is. This is machine search rather than human
search, so it is an analogue of Jaru\v{s}ek and Pel\'anek's predictive metrics
rather than one of them. It nonetheless falls on the same side of their
distinction, and deliberately not on the side of $|V|$, which their logs show
carries almost no signal.

\subsection{Masked diffusion model}

\begin{figure}[htbp]
  \centering
  \includegraphics[width=\textwidth]{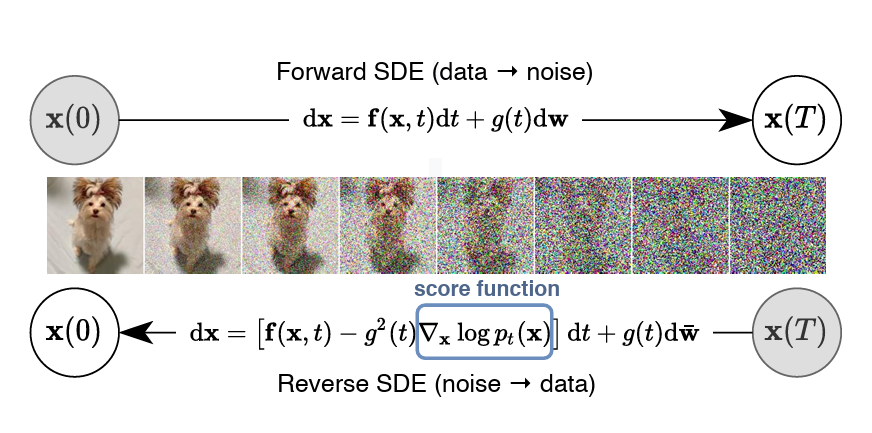}
  \caption{Reproduced from Song et al. \cite{song2021sde}, Figure~1.
  \textbf{Top, forward SDE:} the process that corrupts data $\mathbf{x}(0)$ into
  noise $\mathbf{x}(T)$, shown corrupting a photograph step by step.
  \textbf{Bottom, reverse SDE:} the same process run backwards, turning noise
  back into data. This is made possible only if the \textbf{score function}
  $\nabla_{\mathbf{x}} \log p_t(\mathbf{x})$, boxed in the equation, is known at
  every intermediate step.}
  \label{fig:sde}
\end{figure}

Continuous diffusion models are built on a stochastic differential equation
(SDE) that gradually turns data into noise. Song et al. \cite{song2021sde}
showed this process can be run in reverse in two ways: as a matching
reverse-time SDE, or as a deterministic \emph{probability flow ODE} with the
same marginal distributions, i.e., an ordinary differential equation that a
standard solver can integrate directly. Both routes need the same missing piece:
the \emph{score function}, $\nabla_x \log p_t(x)$, the gradient of the
log-density of the noised data at each step. The score function has no closed
form; it can only be estimated by training a separate network against a
score-matching objective. Every part of this, meaning the SDE, the ODE, the
gradient, the density is defined over a continuous, differentiable space. None of it has
a meaning for discrete data: there is no gradient of a distribution over seven
tile types, and no log-density of a word. Figure~\ref{fig:sde} shows the forward
and reverse processes, and where the score function enters.

The diffusion models that generate images work on \textbf{continuous} data. The
forward process gradually adds Gaussian noise to a picture until nothing is left
but static; the model learns to run that backwards, removing a little noise at a
time. Discrete data however breaks that. A Sokoban cell is a wall, or a floor,
or a box; there is no ``slightly noisy wall'', and nothing sensible halfway
between a wall and a box. Gaussian noise has nothing to act on.

Austin et al. \cite{austin2021d3pm} built a genuinely discrete diffusion process
instead, replacing the SDE with a Markov chain over categorical transition
matrices, with no score function required. However, the resulting training
objective was still a fairly involved categorical ELBO. Shi et al.
\cite{shi2024md4}, in \emph{MD4: Simplified and Generalized Masked Diffusion for
Discrete Data}, showed that for the masking case, specifically, this collapses to
something much plainer: an ordinary cross-entropy loss, computed only at masked
positions and reweighted using the timestep. This project's training algorithm
follows the MD4 formulation directly.

Masked diffusion replaces noising with \emph{hiding}. Corruption means
swapping a token for a special \MASK{} symbol, and the schedule controls how
many tokens are hidden rather than how much noise is added. Generation runs that
backwards: start from a fully masked grid and progressively reveal cells,
predicting what belongs in each. How many are revealed per step follows from the
number of diffusion steps $T$, a design choice we return to in
Section~\ref{sec:training}.

In masked diffusion models commitments are final: once a cell is
unmasked it can never be selected again, and the sampler only ever draws from
cells still marked \MASK{}, so a wall placed at step 3 stays a wall for the rest
of generation. Continuous diffusion has no such rule: every pixel is nudged at
every one of its steps, all the way to the end, so an early bad direction can
still be pulled back later.

\subsection{Contribution}

\begin{figure}[htbp]
  \centering
  \includegraphics[width=0.75\textwidth]{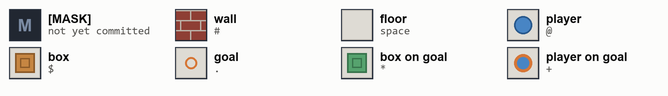}
  \caption{Tile key.}
  \label{fig:legend}
\end{figure}

We train a \textbf{masked diffusion model} to generate Sokoban puzzles, using the
formulation from the MD4 paper \cite{shi2024md4}. The training data is DeepMind's
Boxoban \cite{guez2019boxoban} dataset. Puzzles here are $10\times10$ grids of
tiles rather than sentences of words: each one flattens to \textbf{100 tokens}
over a vocabulary of 7 tile types (\texttt{\#} wall, space floor, \texttt{@}
player, \texttt{\$} box, \texttt{.} goal, \texttt{*} box-on-goal, \texttt{+}
player-on-goal), plus a \MASK{} symbol the model uses but the data never
contains. Figure~\ref{fig:legend} gives the tile key used throughout.

\subsubsection{Solvability emerges without supervision}

Trained only to fill in masked cells, with no solver, reward, or solvability
label in the loop, the model generates puzzles that are \textbf{77.4\%} solvable
unfiltered, rising to \textbf{98.7\%} once failures repairable by deleting a
single interior wall are counted. Here 50,000 puzzles were generated using our
model and every unsolvable puzzle was checked. Counting two-wall repairs as
well, only ${\sim}0.40\%$ of everything generated is genuinely broken.
Interestingly, the culprit walls were committed at a median probability of
\textbf{0.45}, against \textbf{0.93} for the other interior walls of the very
same puzzles.

\subsubsection{Distribution match}

The tile-pattern divergence between generated puzzles and the training corpus
sits on the divergence between \emph{real held-out} puzzles and the same corpus,
at every sample size from 250 to 50,000, where the held-out split runs out. Both
series decay at the same rate, and what separates them is under 4\% of the
divergence itself at every size. Solvability is simply inherited from the
training dataset.

\subsubsection{Temperature trade-off}

Lowering $\tau$ from 1.0 to 0.6 raises solvability by 3.8 points but inflates
average wall count from 69.5 to 73.2 in the temperature sweep, against a corpus
average of 68.6, while cutting median solver effort by 36\%. The default
$\tau = 1.0$ is the setting at which generated puzzles match real wall density.

\section{Method and design choices}

This section lays out the design of the model and the reasoning behind each
choice. We first motivate why a diffusion model, rather than an autoregressive
one, suits Sokoban generation. We then describe the training objective and work
through the three choices that shape it: the loss weighting that keeps every
noise level contributing usefully, the number of diffusion steps, and the noise
schedule. Each is presented not just as a setting but with the argument for why
it takes the value it does.

\subsection{Model choice}

The non-local interdependence between different parts of the puzzle is what
makes it difficult, so an autoregressive generator, which commits to everything
in one fixed order, isn't a good fit for this kind of game. Diffusion does not
work that way. It fills in cells in whatever order the reveal process happens to
land on, and each new cell is conditioned on every cell already decided so far,
wherever in the grid it sits, not on a fixed prefix that always runs in the same
direction. The model might settle a goal in one corner, a wall on the opposite
side, and only later the corridor connecting them, rather than reading the grid
off in one fixed pass. Because a puzzle's actual difficulty comes from exactly
this kind of non-local interaction meaning a decision in one part of the grid
constraining what will work somewhere else entirely, a generator that isn't
locked into a single fixed order is a better structural match for the problem
than one that is.

\subsection{Architecture}

The generator $f_\theta$ is a bidirectional Transformer encoder ($\approx$ 4.9M
parameters): $d=256$, $6$ layers, $8$ heads, feed-forward width $1024$, dropout
$0.1$. Encoder only is used because the task is fill-in-the-blanks over
a grid, not left-to-right generation. The generator maps a masked grid $x_t \in \{0,\dots,7\}^{100}$ and timestep $t$ to per-cell
logits over the $7$ real tiles, $f_\theta(x_t,t) \in \mathbb{R}^{100 \times 7}$
(\MASK{} is an input token only, never predicted). Cell $i$, at grid position $(r_i,c_i)$, is
embedded as
\begin{align*}
  h_i^{(0)} &= E_{\text{tok}}(x_{t,i}) + E_{\text{row}}(r_i)
               + E_{\text{col}}(c_i) + \tau(t), \quad \tau(t)   = \mathrm{MLP}\big(\mathrm{sinusoid}(t)\big) \in \mathbb{R}^{d}.
\end{align*}
Here $E_{\text{tok}}$, $E_{\text{row}}$, and $E_{\text{col}}$ each maps an integer
index to a learned $d$-vector. Separate row/column embeddings give attention
the 2-D grid geometry directly rather than through a flat 1-D index. Moreover,
\begin{align*}
  \mathrm{sinusoid}(t)
    &= \big[\sin(t\omega_0),\cos(t\omega_0),\
            \sin(t\omega_1),\cos(t\omega_1),\ \dots\big], \quad
  \omega_k = 10000^{-2k/d}.
\end{align*}
The sinusoidal form and the base $10000$ are the standard ones introduced for
positional encoding by Vaswani et al. \cite{vaswani2017attention} and carried
over to diffusion timestep embeddings.
Note that the timestep term $\tau(t)$ is added identically to every cell. The
embeddings then feed a stack of 6 standard pre-norm Transformer blocks
(multi-head attention + FFN with residual connections), and a final linear layer
projects each cell to logits over the 7 tiles.

\subsection{Training}
\label{sec:training}

At timestep $t$ each cell is independently replaced by \MASK{} with probability
$1-\alpha_t$, on a linear schedule $\alpha_t = 1 - t/T$ with $T = 100$. At $t=0$
the grid is intact; at $t=T$ it is entirely mask. A training data point then
consists of a sampled puzzle from the training data, a sampled
$t \in \{1,\dots,T\}$, and a masked version of the chosen puzzle via the
scheduler. The model is then trained to recover the original tokens at the masked
positions, where the loss is the cross-entropy weighted by
\begin{equation*}
  w(t) = \min\!\left(\frac{1}{1-\alpha_t},\ w_{\max}\right)
       = \min(T/t,\ w_{\max}),
  \qquad w_{\max}=10 .
\end{equation*}
The loss function is therefore calculated as below.
\begin{equation*}
  \mathcal{L}(\theta) = \mathbb{E}_{x_0, t, m}
    \Big[\, w(t) \cdot \frac{1}{|M|} \sum_{i \in M}
      -\log p_\theta\big( x_0^{(i)} \mid x_t, t \big) \Big] ,
\end{equation*}
where $m$ is the mask draw, with each position hidden independently with
probability $1-\alpha_t$, and $M = \{ i : m_i = 1 \}$ is the set of positions it
hides, where $t \sim \mathrm{Uniform}\{1,\dots,T\}$.
We now explain the three remaining choices: the weight cap, the number of
diffusion steps, and the schedule.

\paragraph{Weights.}
At timestep $t$ only $100 \cdot t/T$ cells are masked in expectation, so at $t=1$
the model is graded on roughly one cell and at $t=T$ on all hundred. Without
reweighting, a step that hides a single cell contributes as much to the gradient
as one that hides the entire grid. The $1/(1-\alpha_t)$ factor which falls out
of the masked-diffusion ELBO restores the per-sequence scale. Note that $t$
never reaches $0$: there $\alpha_t = 1$, so nothing is masked, the loss has no
positions to average over and $w(t)$ is undefined, which is why $t$ is drawn
from $\{1,\dots,T\}$. The ratio $T/t$ is therefore largest at $t=1$, where it
reaches $T$, so with uncapped $w(t)$ a single near-complete grid
would carry $T\times$ the gradient weight of a fully-masked one, and the gradient
becomes dominated by a handful of nearly finished examples. Small $t$ is the
\emph{near-complete} regime, in which a grid has only a handful of cells left to
fill. These are the cells that are critical for solvability. $w_{\max}$ therefore
sets how much the model learns about the phase that determines the global
property it is never trained on.

\paragraph{Number of diffusion steps.}
Write $L$ for the number of cells in a grid, so $L = 100$ here.
Notice that $T=L$ is the unique value where, first, exactly one cell is revealed
per step and, second, no trained timestep is left unused by the sampler. Every
reveal-step must unmask at least one new cell, so sampling always runs
$\min(T, L)$ reveal steps. Choosing $T < L$ forces the sampler to reveal more
than one cell per step; choosing $T > L$ leaves the sampler visiting only $L$ of
the $T$ trained timesteps, so most are never used at inference.

\paragraph{Scheduler.}
As mentioned above, since the loss is computed only at masked positions, a timestep with few masked
tokens carries little information per gradient step, and the uncapped weight
$w^\star(t) \equiv 1/(1-\alpha_t)$ compensates by amplifying it. The schedule
decides how sharply this amplification grows as $t$ approaches its minimum.
Under the linear and cosine schedules, near $t=0$ the masked fraction behaves as
\begin{equation*}
  1-\alpha_t^{\text{linear}} = \frac{t}{T},
  \qquad
  1-\alpha_t^{\text{cosine}} \approx \frac{\pi^2 t^2}{8T^2},
\end{equation*}
i.e., a linear versus a quadratic falloff, so the uncapped weights grow as
\begin{equation*}
  w^\star_{\text{linear}}(t) = \frac{T}{t},
  \qquad
  w^\star_{\text{cosine}}(t) \approx \frac{8T^2}{\pi^2 t^2}.
\end{equation*}
The cosine weight therefore diverges quadratically in $1/t$ rather than
linearly, reaching $w^\star(1) \approx 8106$ against linear's
$w^\star(1) = T = 100$ which is a bounded ceiling equal to the sequence length
itself. Finally, Algorithm~\ref{alg:train} summarizes the training algorithm, following
\cite{shi2024md4}. In Algorithm~\ref{alg:train}, training is shown for a single puzzle; the
implementation vectorizes over a batch of $B$.

\begin{algorithm}[htbp]
\caption{Training step}
\label{alg:train}
\begin{algorithmic}[1]
  \Require Batch of puzzles $\{x_0^{(b)}\}_{b=1}^{B}$, model $f_\theta$
  \State Sample $t^{(b)} \sim \mathrm{Uniform}\{1, \dots, T\}$ for each $b$
  \State Compute $\alpha_t = 1 - t/T$
  \State Sample masks $m_i \sim \mathrm{Bernoulli}(1 - \alpha_t)$ for each position
  \State Create $x_t$: replace $x_0^{(i)}$ with \MASK{} where $m_i = 1$
  \State Forward pass: $\text{logits} = f_\theta(x_t, t)$
  \State Average cross-entropy over the masked positions
  \State Apply weight $w(t) = \min(1/(1 - \alpha_t),\, w_{\max})$
  \State Backpropagate and update $\theta$
\end{algorithmic}
\end{algorithm}

\subsection{Loss and solvability}

The model was trained for 1,000 epochs over the 450,000-puzzle corpus, 292,000
optimizer steps at batch size 1,536, using AdamW (learning rate
$2.45\times10^{-4}$, weight decay 0.01, 125 warmup steps, cosine decay, gradient
clipping at 1.0) in fp16 mixed precision on a single RTX 5070 Ti. Validation
loss is measured every 1,000 steps on Boxoban's held-out split, and solvability
every 50,000 steps by generating 5,000 fresh puzzles from that checkpoint and
running the push solver on each.

\begin{figure}[htbp]
  \centering
  \includegraphics[width=0.85\textwidth]{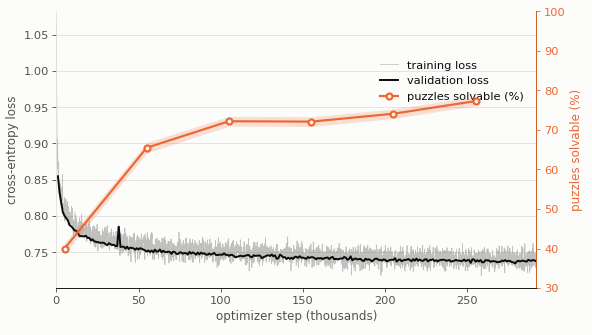}
  \caption{Training and validation loss (left axis) against solvability (right
  axis) over the training run. Solvability is measured on 5,000 freshly
  generated puzzles per checkpoint.}
  \label{fig:training}
\end{figure}

Figure~\ref{fig:training} shows all three. Loss converges early and then stays
flat; solvability keeps climbing to the end of the run. The training objective
is a per-cell reconstruction loss, while solvability is a global property it
never sees, so the two are free to decouple. It is emphasized that a run halted when the loss
flattened would have given up roughly 25 points of solvability. Train and
validation loss also track each other throughout, so the long run is not
overfitting.

\subsection{Distribution match}

To assess whether generated puzzles capture the structural style of the Boxoban
corpus beyond mere solvability, we evaluate the Jensen-Shannon Divergence,
defined as
\begin{equation*}
  \mathrm{JSD}(P \parallel Q) = \tfrac{1}{2}\mathrm{KL}(P \parallel M)
                              + \tfrac{1}{2}\mathrm{KL}(Q \parallel M),
  \qquad M = \tfrac{1}{2}(P + Q),
\end{equation*}
between the empirical $3 \times 3$ sliding-window tile distributions of
generated samples ($Q$) and the 450,000-puzzle training reference ($P$).
Extracting 64 local $3 \times 3$ windows per $10 \times 10$ grid captures
critical structural features such as corridors, corners, and dead ends. We choose JSD
over raw Kullback-Leibler divergence because it is symmetric, bounded in
$[0, 1]$ under base-2 logarithms, and naturally handles unobserved patterns without requiring arbitrary
additive smoothing. Finally, because sample size strongly affects support
coverage and raw scores, every generated JSD score is calibrated directly
against a baseline of real held-out puzzles evaluated at the identical sample
size.

\begin{figure}[htbp]
  \centering
  \includegraphics[width=0.85\textwidth]{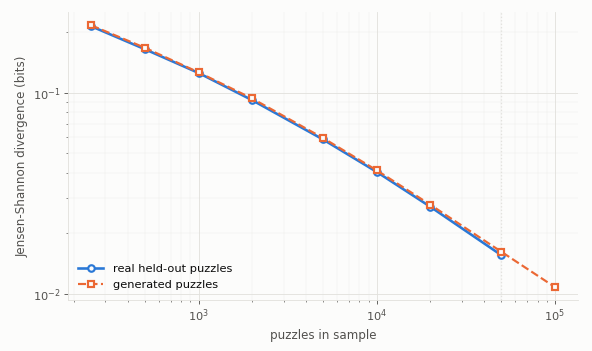}
  \caption{Tile-pattern JSD against the 450,000-puzzle training corpus, plotted
  against sample size, for generated puzzles and for real held-out puzzles
  measured identically. Both series decay as $K^{-0.59}$: the divergence a small
  sample shows is dominated by how few $3\times3$ patterns it can cover, not by
  the source it was drawn from. The held-out curve is therefore the floor, and
  the generated curve sits on it.}
  \label{fig:jsd}
\end{figure}

Figure~\ref{fig:jsd} compares generated puzzles against real ones on the same
measurement: both series show how far a sample of puzzles diverges from the
450,000-puzzle training corpus in its distribution of local $3\times3$ tile
patterns, plotted against how many puzzles went into the sample. The validation
set is Boxoban's held-out split, which comes with DeepMind's dataset and
consists of 50,000 real puzzles that were never shown to the model.

\section{Inference and evaluation}

\begin{figure}[htbp]
  \centering
  \includegraphics[width=\textwidth]{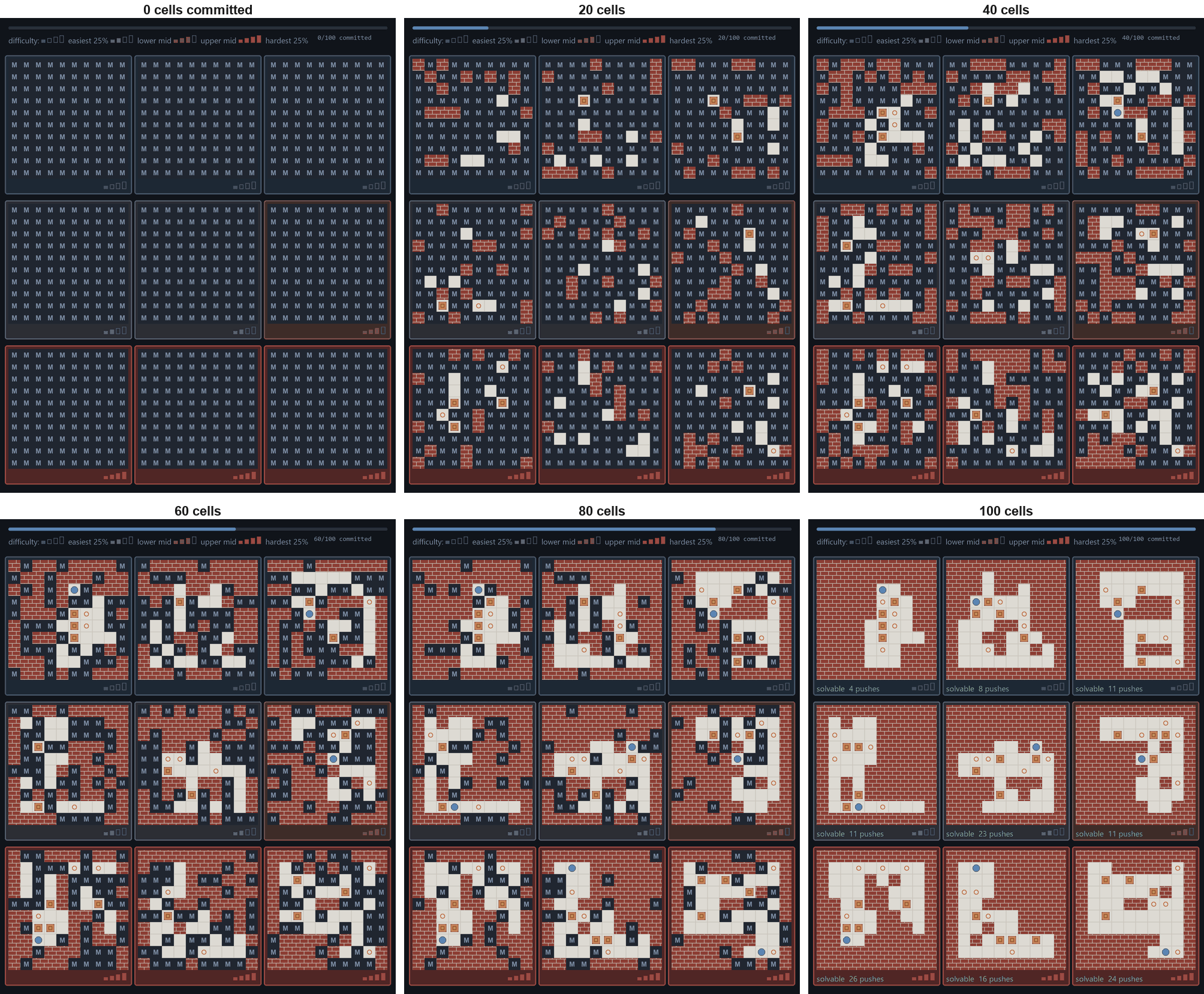}
  \caption{Nine puzzles generated from a fully masked grid, sampled at six
  points across the 100 denoising steps. One cell is committed per step in
  uniformly random order and never revised; cells still showing \texttt{M} are
  masked. Verdicts and push counts appear in the final panel, once every cell is
  committed, and the bars rate each puzzle against the training corpus's
  difficulty quartiles.}
  \label{fig:generate}
\end{figure}

\begin{figure}[htbp]
  \centering
  \includegraphics[width=\textwidth]{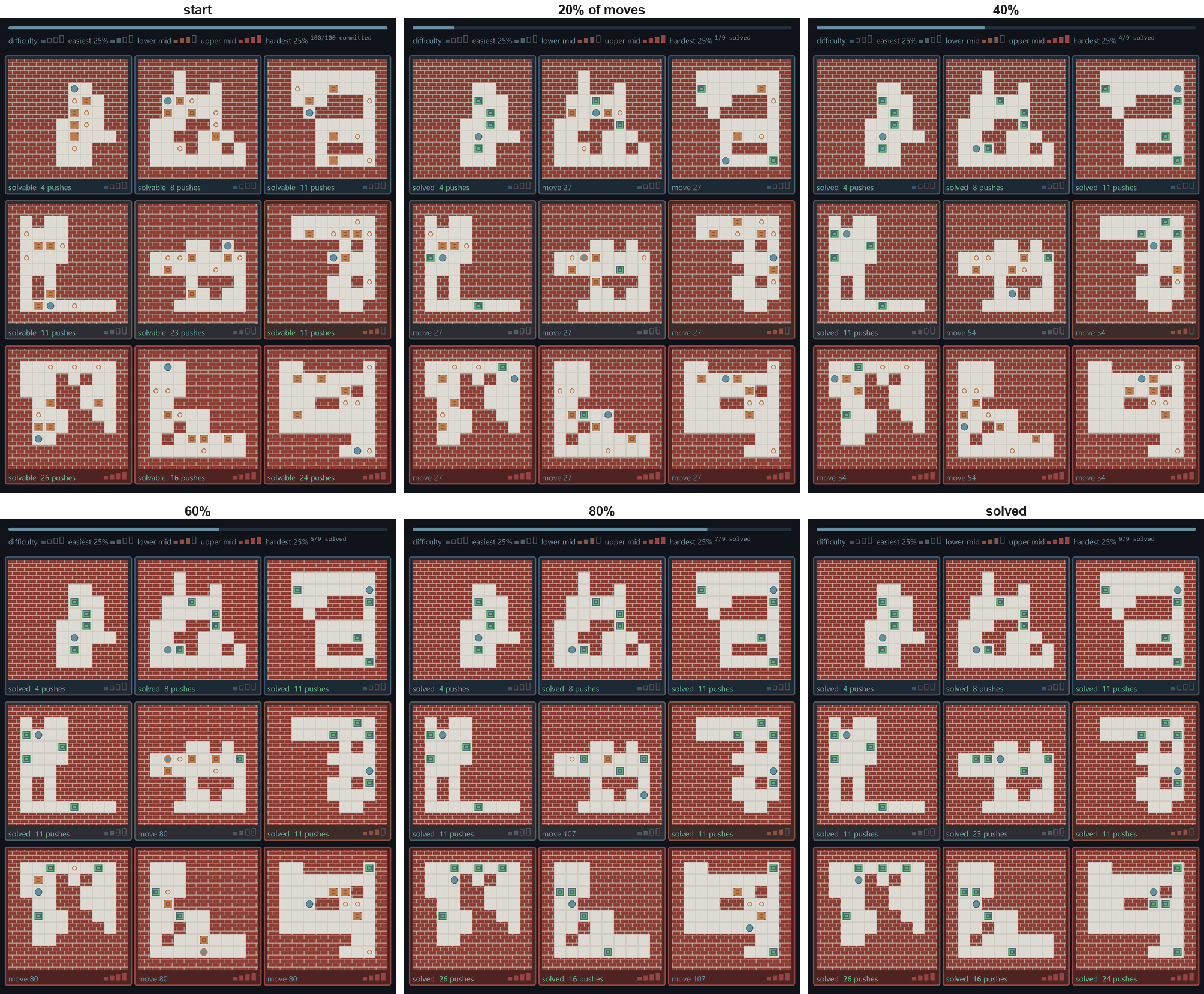}
  \caption{The same nine puzzles, played back under solutions found by the
  push-based solver, sampled at six points through the playback. The solver is
  run only for evaluation and plays no part in generation.}
  \label{fig:solve}
\end{figure}

This section reports the evaluation results for the trained model: solvability,
one-wall repairability, memorization, and the effect of sampling temperature. We
begin with sample puzzles (Figure~\ref{fig:generate}), rated for difficulty by
the push-based search described earlier. Below are the solutions to those
puzzles (Figure~\ref{fig:solve}).

\subsection{Sampling algorithm}
\label{sec:sampling}

The generation process reverses corruption by starting with a 100-cell \MASK{}
grid and unmasking one cell at a time across 100 steps. At each step, the model
predicts probability distributions over all seven tile types across the entire
grid simultaneously. A candidate tile is sampled from each distribution---rather
than selected via argmax---and exactly one uncommitted cell is chosen uniformly
at random to be fixed. Discarding the remaining predictions and recomputing them
from scratch on each step ensures the model chooses what tile to place while a
uniform sampler determines which cell comes next. By conditioning each step on
one additional finalized cell, the system bypasses having to model the full
100-cell joint distribution directly. In other words, the following holds
\begin{equation*}
  P(c_1, \dots, c_{100}) = \mathbb{E}_{\sigma}
    \left[\prod_{k=1}^{100}
      P\big(c_{\sigma(k)} \mid c_{\sigma(1)}, \dots, c_{\sigma(k-1)}\big)\right],
\end{equation*}
where $\sigma$ is the random order in which cells happen to be revealed. Because
training masks an arbitrary subset rather than a prefix, the model learns
$p_\theta(c_i \mid c_S)$ for conditioning sets $S$ of every size and shape, which
is precisely what allows any reveal order to be used at sampling time.

Selecting cells uniformly also avoids the pitfalls of confidence-based ordering:
because walls are the easy, high-confidence majority class, committing
high-confidence cells first biases subsequent predictions toward even more walls,
inflating the average wall count in that comparison from 69.5, which matches the
corpus average of 68.6, up to 81.5. Algorithm~\ref{alg:sample} provides the
inference algorithm.

\begin{algorithm}[htbp]
\caption{Sampling}
\label{alg:sample}
\begin{algorithmic}[1]
  \Require Trained model $f_\theta$, $T = 100$ steps, temperature $\tau = 1.0$
  \State Initialize $x \leftarrow \MASK^{100}$
  \For{$\text{step} = 0, \dots, T-1$}
    \State Compute $t = T - \text{step}$
    \State Forward pass:
           $p = \mathrm{softmax}\big(f_\theta(x, t) / \tau\big)$ for every cell
    \State Draw a candidate tile for every cell:
           $\hat{x}_i \sim \mathrm{Categorical}(p_i)$
    \State Collect the still-masked cells $M = \{ i : x_i = \MASK \}$
    \State Decide how many to reveal:
           $n = \max\big(\lceil |M| / (T - \text{step}) \rceil, 1\big)$,
           which is $1$ here
    \State Choose which cells to reveal: pick $n$ of the masked cells in $M$
           at random, all equally likely
    \State Commit the values sampled in line 5 at those cells
           (never revised afterwards)
  \EndFor
  \State \Return $x$
\end{algorithmic}
\end{algorithm}

\subsection{One-wall fixes}

\begin{figure}[htbp]
  \centering
  \includegraphics[width=0.75\textwidth]{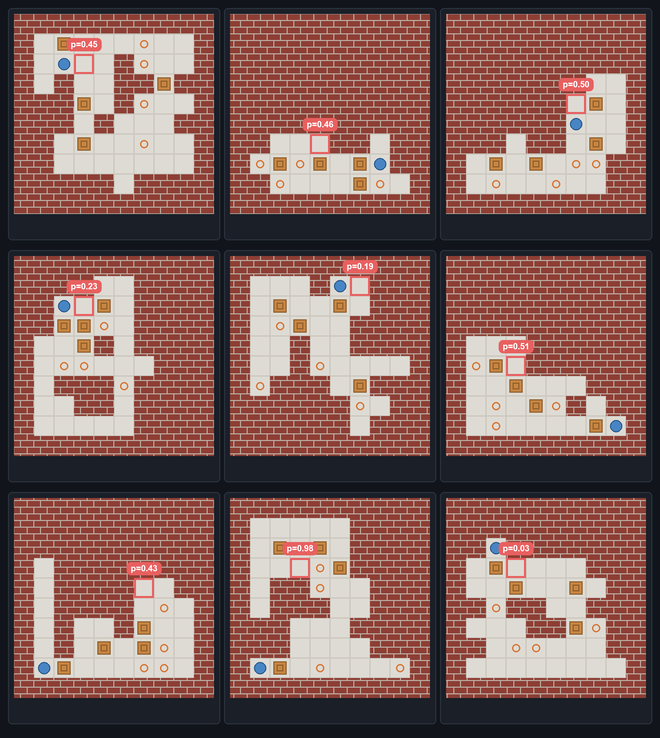}
  \caption{Nine unsolvable puzzles, each repaired by deleting one interior wall.
  The red outline marks the removed cell and the chip gives the probability the
  model assigned it when it was committed during generation. Measured on 11,288
  unsolvable puzzles from a 50,000-sample run.}
  \label{fig:wallfix}
\end{figure}

When a generated puzzle is unsolvable, the failure is usually shallow rather than
structural: \textbf{94.5\%} of unsolvable puzzles become solvable by deleting a
single interior wall, which lifts effective solvability from 77.4\% to
\textbf{98.7\%}. Each puzzle in Figure~\ref{fig:wallfix} displays the
probability the model assigned that wall at the moment it committed it during
generation. These probabilities have a median of \textbf{0.45}, against
\textbf{0.93} for the other interior walls of the very same puzzles.

\subsection{Memorization}

A generator that reproduced its training data would score well on every metric
above while being worthless. For each generated puzzle we measure the Hamming
distance to its nearest neighbour among the 450,000 training puzzles. This is
equal to the number of the 100 cells on which the two grids differ, so 0 is an
exact reproduction.

The player's position is canonicalised away before comparing. The worker moves
freely within its reachable region, and the push solver collapses those positions
into a single state for exactly that reason, so two grids differing only in where
the worker stands are the same puzzle. Counting that as a difference would
undercount duplicates.

On its own, a distance of say 12 means little: every Boxoban puzzle shares the
same wall border and similar density, so unrelated puzzles already might agree on
most cells. We use the 50,000 held-out puzzles as the reference. They come from
the same distribution but were never shown to the model, so they cannot have been
memorised. We observe that they produce the same curve once compared against the
training corpus.

\begin{figure}[htbp]
  \centering
  \includegraphics[width=0.85\textwidth]{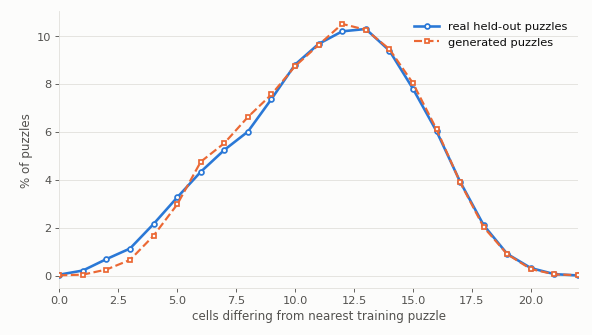}
  \caption{Nearest-neighbour Hamming distance to the training corpus, player
  position canonicalised away. \textbf{Blue:} for each of 50,000 held-out
  puzzles, the distance to its nearest neighbour among the 450,000 training
  puzzles. \textbf{Orange:} the same measurement for 50,000 generated puzzles.
  The blue series is the reference: it is what a generator that memorised nothing
  would score. Mass piled up near zero in the orange series would be
  memorisation; there is none.}
  \label{fig:hamming}
\end{figure}

Over 50,000 generated puzzles and 50,000 held-out puzzles, the two distributions
are near-identical: median distance 12 for both, mean 11.43 against 11.33. Exact
reproductions are rarer in the generated set than in the real one, 5 against 19,
as is close agreement, 5.6\% within 5 cells against 7.5\%. The model is therefore
not copying; if anything it lands slightly further from the training corpus than
genuine puzzles do. Figure~\ref{fig:hamming} plots both distributions.

\subsection{Sampling temperature}

\begin{figure}[htbp]
  \centering
  \includegraphics[width=0.85\textwidth]{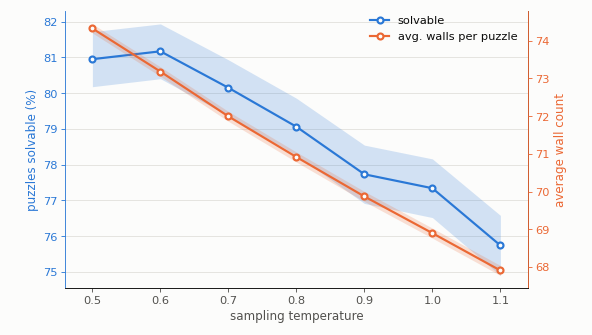}
  \caption{Solvability and average wall count against sampling temperature,
  10,000 samples per setting on the $T = 100$ checkpoint. Shaded bands are 95\%
  confidence intervals. The training corpus averages 68.6 walls per puzzle;
  sampling at $\tau = 1.0$ matches it, and every lower setting exceeds it.}
  \label{fig:temperature}
\end{figure}

Temperature $\tau$ rescales the logits before each cell is drawn. Lowering it
sharpens the distribution toward the model's top choice; raising it flattens it.
Figure~\ref{fig:temperature} describes the impact of temperature on wall count
and solvability.

\section{Conclusion}

A masked diffusion model trained purely on tile completion, with no solver,
reward, or solvability label at any point in training, sampling, or filtering,
generates Sokoban puzzles that are 77.4\% solvable unfiltered. Counting failures
repairable by deleting a single interior wall raises this to 98.7\%, and
counting two-wall repairs leaves only ${\sim}0.40\%$ of generated puzzles
genuinely broken. Solvability is a global property, PSPACE-complete to decide
and with no short certificate to check, yet it follows here from an objective
that only ever asks the model to fill in masked cells.

Two checks argue that this is inherited rather than trivial or copied. The
tile-pattern divergence between generated puzzles and the training corpus sits
on the divergence between real held-out puzzles and that same corpus, at every
sample size from 250 to 50,000 and to within 4\% of the divergence itself, so
the model reproduces the corpus structure rather than retreating to some easy
subset of it. And nearest-neighbour Hamming distance rules out memorisation:
generated and held-out puzzles give a median distance of 12 apiece, with exact
reproductions rarer among generated puzzles than among real ones, 5 against 19.
What the model has learned is the training distribution, and solvability comes
with it.

Two observations are worth carrying to other work of this kind. First, loss and
solvability decouple: validation loss converges early and then stays flat while
solvability keeps climbing to the end of the run, so a run halted when the loss
flattened would have forfeited roughly 25 points of solvability. A per-cell
reconstruction loss is not a proxy for a global structural property, and should
not be used as a stopping criterion for one. Second, the failures that remain
are shallow and the model half-signals them itself: the walls whose removal
repairs a puzzle were committed at a median probability of 0.45, against 0.93
for the other interior walls of the very same puzzles. Last but not least, sampling temperature
trades along the same axis, with $\tau = 1.0$ the setting at which generated
wall density matches the corpus.

\bibliographystyle{plain}
\bibliography{refs}

\end{document}